%% file: main.tex
\documentclass[10pt,journal,compsoc]{IEEEtran}
\usepackage{times}
\usepackage{epsfig}
\usepackage{graphicx}
\usepackage{amsmath}
\usepackage{amssymb}
\usepackage{pifont}
\usepackage{xcolor}
\usepackage{framed}
\usepackage{amssymb}
\usepackage{latexsym}
\usepackage{epsfig}
\usepackage{float}
\usepackage{booktabs,multirow, multicol}
\usepackage{rotating}
\usepackage{epstopdf}
\usepackage{graphicx}
\usepackage{hyperref}

\usepackage{subcaption}

\usepackage[normalem]{ulem}
\usepackage{hhline}
\useunder{\uline}{\ul}{}

\usepackage{array, longtable}
\usepackage{wrapfig}
\usepackage{enumitem}
\usepackage{tabularx}
\usepackage{url}
\usepackage[altpo]{backnaur}

\usepackage{epsfig}
\usepackage{graphicx}
\usepackage{amsmath}
\usepackage{amssymb}
\usepackage{gensymb}
\usepackage{footmisc}
\usepackage{pifont}
\PassOptionsToPackage{table}{xcolor}
\usepackage{array}
\usepackage{url}

\usepackage{collcell}
\usepackage{xcolor}
\usepackage{colortbl}

\usepackage{listings}
\usepackage{xcolor}
\usepackage{fancyvrb}

\definecolor{Gray}{rgb}{0.96, 0.76, 0.76}
\newcommand*{\MinNumber}{-0.01}%

\newcommand{\AG}[1]{%

        \ifdim #1 pt > \MinNumber pt
            \textcolor{black}{#1}

        \fi

}

\ifCLASSOPTIONcompsoc
  \usepackage[nocompress]{cite}
\else
  \usepackage{cite}
\fi

\ifCLASSINFOpdf
\else
\fi

\newcommand{\model}{EdgeFace{} }

\begin{document}

\title{\model\!: Efficient Face Recognition Model \\ for Edge Devices}

\author{Anjith~George,~\IEEEmembership{Member,~IEEE,}
        Christophe~Ecabert,~\IEEEmembership{Member,~IEEE,}
        Hatef~Otroshi~Shahreza,~\IEEEmembership{Graduate~Student~Member,~IEEE,}
        Ketan~Kotwal,~\IEEEmembership{Senior~Member,~IEEE,}
        S\'ebastien~Marcel,~\IEEEmembership{Senior~Member,~IEEE}
\IEEEcompsocitemizethanks{\IEEEcompsocthanksitem All authors are with Idiap Research Institute, Martigny, Switzerland. Hatef Otroshi Shahreza is also affiliated with \'{E}cole Polytechnique F\'{e}d\'{e}rale de Lausanne (EPFL), Lausanne, Switzerland,  and S\'{e}bastien Marcel is also affiliated with Universit\'{e} de Lausanne (UNIL), Lausanne, Switzerland.
E-mail: \{anjith.george, christophe.ecabert, hatef.otroshi, ketan.kotwal, sebastien.marcel\}@idiap.ch 
}
\thanks{Manuscript received August XX, 2023.}}

\markboth{Journal of \LaTeX\ Class Files,~Vol.~14, No.~8, August~2015}%
{Shell \MakeLowercase{\textit{et al.}}: Bare Demo of IEEEtran.cls for Biometrics Council Journals}

\IEEEtitleabstractindextext{%
\begin{abstract}

In this paper, we present \model\!- a lightweight and efficient face recognition network inspired by the hybrid architecture of EdgeNeXt. By effectively combining the strengths of both CNN and Transformer models, and a low rank linear layer, \model achieves excellent face recognition performance optimized for edge devices. The proposed \model network not only maintains low computational costs and compact storage, but also achieves high face recognition accuracy, making it suitable for deployment on edge devices. The proposed \model model achieved the top ranking among models with fewer than 2M parameters in the IJCB 2023 Efficient Face Recognition Competition. Extensive experiments on challenging benchmark face datasets demonstrate the effectiveness and efficiency of \model in comparison to state-of-the-art lightweight models and deep face recognition models. Our \model model with 1.77M parameters achieves state of the art results on LFW (99.73\%), IJB-B (92.67\%), and IJB-C (94.85\%), outperforming other efficient models with larger computational complexities. The code to replicate the experiments will be made available publicly.
\end{abstract}

\begin{IEEEkeywords}
Efficient Face Recognition, Edge Devices, Face Recognition
\end{IEEEkeywords}}

\maketitle

\IEEEdisplaynontitleabstractindextext

%
\IEEEpeerreviewmaketitle

\section{Introduction}
\label{sec:intro}

\IEEEPARstart{F}ace recognition has become an increasingly active research field, 
 achieving significant recognition accuracy by 
leveraging breakthroughs in various computer vision tasks through the development of deep neural networks~\cite{he2016deep,tan2019efficientnet,dosovitskiy2020image} and margin-based loss functions ~\cite{deng2019arcface,wang2018cosface,huang2020curricularface,meng2021magface,boutros2022elasticface,kim2022adaface,terhorst2023qmagface}. 
In spite of remarkable improvements in recognition accuracy, state-of-the-art face recognition models typically involve a deep neural network with a high number of parameters (which requires a large memory) and considerable computational complexity.
Considering memory and computational requirements, it is challenging to deploy state-of-the-art face recognition models on resource-constrained devices, such as mobile platforms, robots, embedded systems, etc. 

To address the issue of memory and computational complexity of state-of-the-art deep neural networks, researchers have been focusing on designing lightweight and efficient neural networks for computer vision tasks that can achieve a better trade-off between recognition accuracy, on one side, and required memory and computational resources, on the other side~\cite{deng2019lightweight,martinez2021benchmarking,boutros2021mixfacenets,boutros2022pocketnet}.
Recently, some works have attempted to utilize lightweight convolutional neural network (CNN) architectures, such as MobileNets~\cite{howard2017mobilenets,sandler2018mobilenetv2}, ShuffleNet~\cite{zhang2018shufflenet,ma2018shufflenet}, VarGNet~\cite{zhang2019vargnet}, and MixNets~\cite{tan2019mixconv},  for face recognition tasks~\cite{chen2018mobilefacenets,martindez2019shufflefacenet,boutros2021mixfacenets,yan2019vargfacenet},  reducing model parameters as well as computational complexity and meanwhile maintaining high levels of accuracy.  
However, with the recent emergence of vision transformers (ViTs) \cite{khan2022transformers} and their ability in modeling global interactions between pixels, there is an opportunity to further improve the efficiency and performance of face recognition models by leveraging both CNNs and ViTs capabilities.

In this paper, we present \model\!, a novel lightweight face recognition model inspired by the \textit{hybrid} architecture of EdgeNeXt~\cite{maaz2023edgenext}. 
We adapt the  EdgeNeXt architecture for face recognition and also introduce a Low Rank Linear (LoRaLin) module to further reduce the computation in linear layers while providing a minimal compromise to the performance of the network. LoRaLin replaces a high-rank matrix in a fully connected layer with two lower-rank matrices, and therefore reduces the number of parameters and required number of multiply adds (MAdds).
\model effectively combines the advantages of both CNNs and ViTs, utilizing a split depth-wise transpose attention (STDA) encoder to process input tensors and encode multi-scale facial features, while maintaining low computational costs and compact storage requirements. Through extensive experimentation on challenging benchmark face datasets, including LFW, CA-LFW, CP-LFW, CFP-FP, AgeDB-30, IJB-B, and IJB-C, we demonstrate the effectiveness and efficiency of \model in comparison to state-of-the-art lightweight models and deep face recognition models, showing its potential for deployment on resource-constrained edge devices. Variants of the proposed \model model achieved the top ranking among models with fewer than 2M parameters in the IJCB 2023 Efficient Face Recognition Competition \cite{kolf2023efar}.
The main contributions of our work can be summarized as follows:

\begin{itemize}
    \item We propose an efficient lightweight face recognition network, called \model\!, based on a {hybrid} network architecture that leverages CNN and ViT capabilities. 
    We adapt the hybrid network architecture of EdgeNeXt for the face recognition task. To the best of our knowledge, this is the first work that uses a hybrid CNN-transformer for efficient face recognition. 
    \item We introduce a Low Rank Linear (LoRaLin) module to further reduce the computation in linear layers while providing a minimal compromise to the performance of the network. LoRaLin module replaces a high-rank matrix in a fully connected layer with two lower-rank matrices, and therefore reduces  the number of parameters and required computations.
    \item We provide extensive experimental results on various challenging face recognition datasets, demonstrating the superior performance of \model in comparison to existing lightweight models. Our experiments also highlight the model's robustness under different conditions, such as pose variations, illumination changes, and occlusions.
\end{itemize}

The source code will be made available publicly \footnote{\url{https://gitlab.idiap.ch/bob/bob.paper.tbiom2023_edgeface}}.

The remainder of this paper is organized as follows. Section~\ref{sec:related_work} provides a brief overview of related works, discussing the limitations of existing lightweight face recognition models and the potential advantages of hybrid architectures. Section~\ref{sec:approach} presents a detailed description of the proposed \model model and the overall hybrid architecture. Section~\ref{sec:experiment} outlines the experimental setup, datasets, and evaluation metrics used to assess the performance of \model, followed by a comprehensive analysis of the results in Section~\ref{sec:discussions}. Finally, Section~\ref{sec:conclusions} concludes the paper and outlines potential future directions for this research.

\section{Related Work} 
\label{sec:related_work}

Over the past decade, face recognition (FR) has been regarded as one of the most
prominent and widely deployed applications of deep learning. However, as the
handheld mobile devices and edge computing became prevalent, the researchers
directed efforts towards developing lightweight FR models without compromising
their accuracy. Since then the design of lightweight CNNs has emerged as an
active research area in general machine learning; and various lightweight
architectures have been developed for common ML applications, such as object
recognition. Inspired by the success of smaller ML models, the biometric
community has adapted some of these lightweight architectures for FR tasks. We
begin with an overview of some of the general-purpose lightweight architectures
and then discuss advances in lightweight and efficient FR.

\subsection{Lightweight Neural Architectures}

With the introduction of MobileNets
\cite{howard2017mobilenets,sandler2018mobilenetv2}, the use of depth-wise
separable convolutions became a major factor in further improving the aspects of
model parameters and FLOPs. The family of MobileNet architectures splits the
typical convolution layer into depthwise (or channelwise) convolutions, followed
by pointwise convolutions. These architectures are small and low-latency, and
thus well-suited for deployment on handheld or embedded applications.
Iandola~\textit{et al.} proposed SqueezeNet architecture that achieved
state-of-the-art recognition accuracy then, despite having $50\times$ fewer
parameters than the contemporary models~\cite{iandola2016squeezenet}. The
SqueezeNet architecture uses a `fire` module as its core block. This module
consists of squeeze and expand operations that are performed by $1\times 1$
convolution filters and a mix of $1 \times 1$ and $3 \times 3$ filters,
respectively. Zhang~\textit{et al.} constructed ShuffleNet architectures-- that employ
pointwise group convolutions and channel shuffling-- for efficient processing on
mobile devices~\cite{zhang2018shufflenet}. Several variants of ShuffleNet can
be defined by applying a scale factor to the number of channels.

The vanilla depth-wise convolutions are extended by incorporating multiple
kernel sizes in a single convolution to design
MixConvNets~\cite{tan2019mixconv}. Kernels with different sizes simplify
capturing multiple patterns from the input using a layer of depthwise
convolutions. By replacing depthwise convolutions in MobileNets (v1 and v2) with
the MixConv feature maps, the MixConvNets are able to achieve better accuracy
with a slight drop in the number of parameters and FLOPs compared to the
baseline MobileNet architectures. The ShiftNet, proposed in \cite{wu2018shift},
is a family of CNNs with a
`shift' block that is a FLOP-free alternative to expensive convolution
operation. The shift block, along with pointwise convolutions, efficiently mixes
spatial information across channels and helps attain a competitive performance for
several tasks, such as classification and style transfer.

In~\cite{zhang2019vargnet}, variable group convolutions were proposed specifically for
deploying neural networks on embedded systems such as FPGA or ASIC. The
variable group operations are targeted at balancing the computational complexity
inside a network block (primarily depthwise separable convolutions).
A series of linear transformations is used to generate \textit{ghost} feature
maps with cheap computation in~\cite{han2020ghostnet}. By stacking such Ghost
modules, Han~\textit{et al}. proposed a GhostNet architecture that requires
fewer parameters and computations compared to the architectures using vanilla
convolutional networks. Different variants of GhostNets can be generated by
controlling hyperparameters related to the number of intrinsic feature maps and
kernel size.

Vision Transformer (ViT) architectures \cite{dosovitskiy2020image} have achieved
excellent results for various recognition tasks, but their high computational
costs have restricted the usage of vision transformers in a low-resource
environment. To address this shortcoming, Chen \textit{et al.} combined local
processing in CNNs (such as MobileNet) and global interaction in transformers to
design a new architecture, Mobile-Former \cite{chen2022mobile}. Mehta and
Rastegari introduced MobileViT architecture, based on local-global image context
fusion, to build a lightweight and low latency network for general vision tasks
\cite{mehta2021mobilevit}. While both aforementioned architectures attempt to
leverage the benefits of CNNs and transformers for vision-classification tasks, the
computational complexity of their MHA (multi-head attention) blocks still
remains a bottleneck for the inference time on edge devices. 

\subsection{Lightweight FR Architectures}

MobileFaceNets are a family of efficient CNN models, based on MobileNet
architecture~\cite{howard2017mobilenets,sandler2018mobilenetv2}, designed for
real-time face verification tasks \cite{chen2018mobilefacenets}. It achieved
99.55\% accuracy on LFW while using less than 1M parameters. The Efficient
Lightweight Attention Networks (ELANet) consist of inverted residual blocks
(similar to MobileNetV2), and additionally, employ concurrent channel- and
spatial-level attention mechanisms \cite{zhang2022efficient}. The ELANets have
nearly 1M parameters, and achieve state-of-the-art performance across multiple
datasets.

The MixConv concept~\cite{tan2019mixconv} was used to develop MixFaceNet
networks for lightweight FR~\cite{boutros2021mixfacenets}. The XS
configuration of MixFaceNet has been reported to exhibit high recognition
performance with as low as 1M parameters.

The FR model using ShiftNet architecture~\cite{wu2018shift} with 0.78M
parameters, called ShiftFaceNet, achieves a comparable performance to that of
FaceNet in terms of recognition accuracy. Duong \textit{et al.} considered
faster downsampling of spatial data/ feature maps and bottleneck residual blocks
towards developing lightweight FR models \cite{duong2019mobiface}. Their
MobiFace and Flipped-MobiFace models provide more than 99.70\% accurate results
on LFW dataset. Inspired from the ShuffleNetV2 \cite{ma2018shufflenet}, the
family of lightweight models, referred to as ShuffleFaceNet, for FR was proposed
in \cite{martindez2019shufflefacenet}. The number of parameters in these models
vary from 0.5M to 4.5M while verification accuracies of higher than 99.20\% have
been reported for LFW dataset. Another family of lightweight architectures,
ConvFaceNeXt \cite{hoo2022convfacenext} uses enhanced version of ConvNeXt blocks
and different downsampling strategies to reduce the number of parameters as well
as FLOPs. With about 1M parameters and nearly 400M FLOPs, ConvFaceNeXt networks
achieve a comparable performance in FR. 

In \cite{boutros2022pocketnet}, neural architecture search (NAS) was used to
automatically design an efficient network- PocketNet, for face recognition.
The PocketNet architecture was learnt using differential architecture search
(DARTS) algorithm on CASIA-WebFace dataset \cite{yi2014learning}. The training
of this network also comprises a multi-step knowledge distillation (KD).
Another approach involving KD for training a face recognition network was
employed in \cite{yan2019vargfacenet}. Their model uses variable group
convolutions to handle the unbalance of computational intensity. The
corresponding model, called VarGFaceNet, was the winner of Lightweight Face
Recognition (LFR) challenge at ICCV 2019 \cite{deng2019lightweight}. In
\cite{shahreza2023synthdistill}, authors introduced a distillation framework
called SynthDistill, and shown that lightweight models can be trained
using synthetic data in an online distillation framework.

Recently, Alansari \textit{et al.} proposed GhostFaceNets (multiple
configurations) that exploit redundancy in convolutional layers to create
compact networks \cite{alansari2023ghostfacenets}. In these modules, a certain
fixed percentage of the convolutional feature maps are generated using depthwise
convolutions that are computationally inexpensive. With configurable
hyperparameters, GhostFaceNets can be designed to contain as low as 61M FLOPs
with nominal reduction in their recognition performance.

In \cite{shahreza2023synthdistill}, lightweight networks (called TinyFaR), based
on TinyNet structure~\cite{han2020model}, were suggested and were trained by KD
from pretrained FR model using synthetic data. For training the lightweight
network within the KD framework, a face generator model was used to generate
synthetic face images, and the lightweight network (as a student) was optimized
to generate the same embedding as the pretrained face recognition model (as a
teacher). This work used dynamic sampling to help the student network focus on
difficult images while exploring newer synthetic images.

\section{Proposed \model Architecture}
\label{sec:approach}
\begin{figure*}[htb!]
    \centering
    \includegraphics[width=0.95\textwidth]{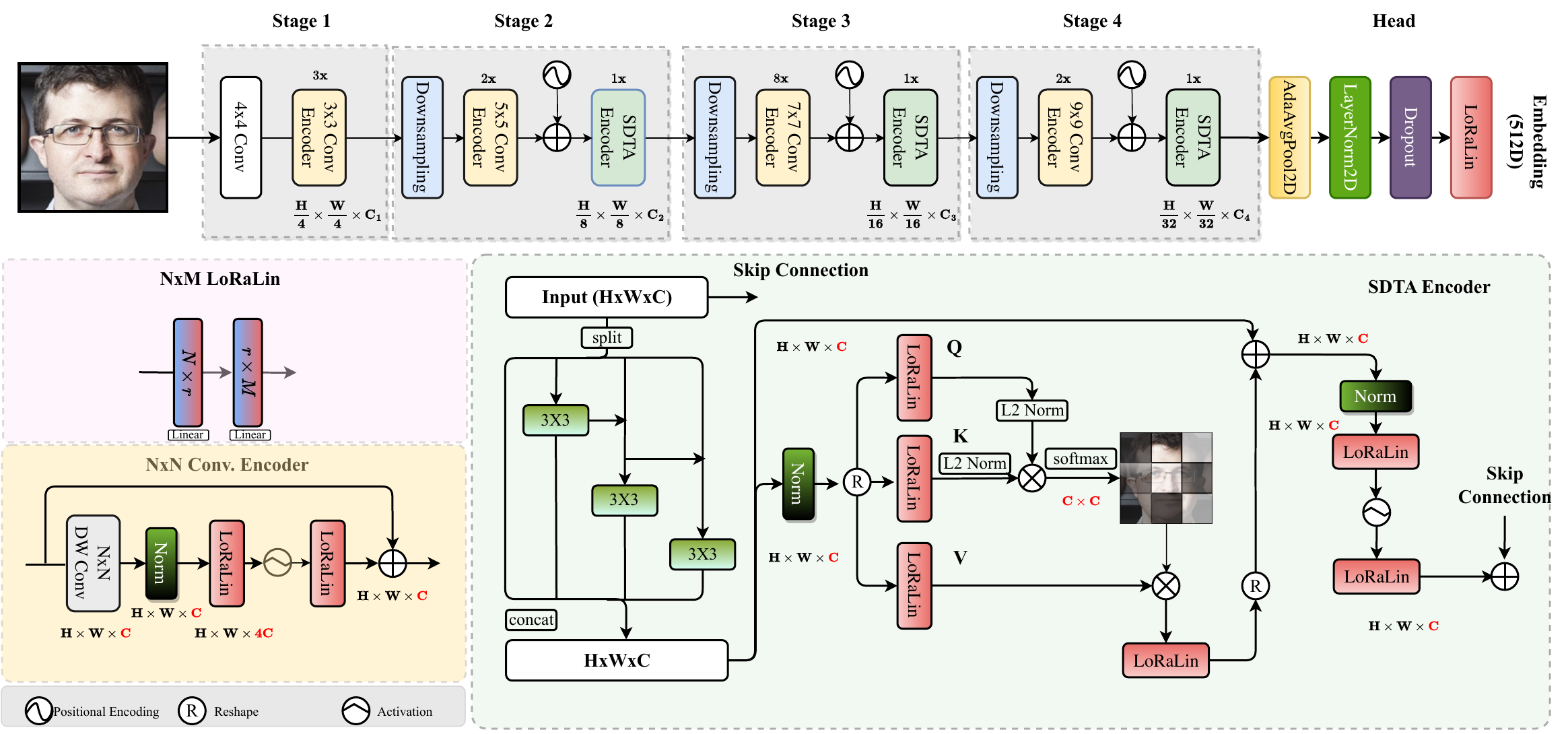} 
    \caption{A schematic diagram of the proposed \model Face Recognition model. The image is adapted from the EdgeNeXt \cite{maaz2023edgenext} model to show the additional elements added to convert it to a face recognition network. Specifically, we introduce \emph{LoRaLin} layers and add a head to obtain the 512-dimensional embeddings.}
    \label{fig:model}
\end{figure*}

In this section, we describe the detailed architecture of the \model FR model. While most of the works on efficient face recognition networks focus on variants of CNNs, they have two primary constraints due to their convolution operations. Firstly, they possess a local receptive field, making it challenging for them to represent global context. Secondly, the weights learned by CNNs remain static during inference, limiting their adaptability to different input content. Transformers and CNN-Transformer hybrids attempt to address these limitations, despite their higher computational cost. In this work, we introduce a lightweight FR model inspired from the CNN-Transformer hybrid architecture of the EdgeNeXt model introduced in \cite{maaz2023edgenext}. We adapt this model to make it suitable for the face recognition task with a focus on reducing the parameters and FLOPs.

\subsection{\model Face Recognition Model}

The primary focus of this work is to design an efficient network tailored for face recognition on edge devices. Towards this goal, we extend the EdgeNeXt \cite{maaz2023edgenext} architecture for face recognition.  First, we try to reduce the parameters and FLOPs of the model further by replacing the Linear layers in the EdgeNeXt network with the newly introduced low rank \emph{LoRaLin} layers. In addition, we add a classification head composed of Adaptive Average Pooling and layer norms, followed by a \emph{LoRaLin} layer outputting a 512-dimensional representation. The input resolution required for the model is adjusted to be $112 \times 112$. To optimally train this adapted model for face recognition, we employ end-to-end training in conjunction with a CosFace \cite{wang2018cosface} classification head. Figure \ref{fig:model} provides a schematic representation of the updated \model face recognition model. First, we  detail the architecture of the EdgeNeXt model designed for image classification, followed by our new additions to make it an efficient face recognition network.

\begin{table}[tbh!]
\centering
\caption{Model Layer Structure, and output dimensions of intermediate layers for different variants of \model}
\resizebox{\columnwidth}{!}{
\begin{tabular}{lcccc}
\toprule
\multirow{2}{*}{\textbf{Layer (depth)}} & \multicolumn{1}{l}{\multirow{2}{*}{\textbf{O/P size}}} & \multicolumn{3}{c}{\textbf{Channels}}                                         \\ \cmidrule(l){3-5} 
                               & \multicolumn{1}{l}{}                          & \multicolumn{1}{c|}{\textbf{SMALL}} & \multicolumn{1}{c|}{\textbf{X-SMALL}} & \textbf{XX-SMALL} \\ \midrule

Sequential 2-1                 & 28$\times$28                                  & 48     & 32       & 24       \\
Conv2d 3-1                     & 28$\times$28                                  & 48     & 32       & 24       \\
LayerNorm2d 3-2                & 28$\times$28                                  & 48     & 32       & 24       \\
EdgeFace-Stage 3-3             & 28$\times$28                                  & 48     & 32       & 24       \\
EdgeFace-Stage 3-4             & 14$\times$14                                  & 96     & 64       & 48       \\
EdgeFace-Stage 3-5             & 7$\times$7                                    & 160    & 100      & 88       \\
EdgeFace-Stage 3-6             & 3$\times$3                                    & 304    & 192      & 168      \\
AdaptiveAvgPool2d 4-9          & 1$\times$1                                    & 304    & 192      & 168      \\
LayerNorm2d 3-8                & 1$\times$1                                    & 304    & 192      & 168      \\
Flatten 3-9                    & -                                           & 304    & 192      & 168      \\
Dropout 3-10                   & -                                           & 304    & 192      & 168      \\
Linear 3-11                    & -                                           & 512    & 512      & 512      \\ \midrule

\textbf{MPARAMS} & - &5.44 &2.24 &1.24 \\
\textbf{MFLOPS} & - &461.7 &196.9 &94.7 \\
\bottomrule
\end{tabular}
}
\label{tab:layers}
\end{table}

\subsection{EdgeNeXt Architecture}

The EdgeNeXt Architecture \cite{maaz2023edgenext} is a lightweight hybrid design that combines the merits of Transformers \cite{vaswani2017attention,dosovitskiy2020image} and Convolutional Neural Networks (CNNs) for low-powered edge devices. EdgeNeXt models with a smaller number of parameters, model size and multiply-adds (MAdds) and outperforms models such as MobileViT \cite{mehta2021mobilevit} and EdgeFormer \cite{EdgeFormer} in image recognition performance. The EdgeNeXt model builds on ConvNeXt~\cite{ConvNeXt} and introduces a new component known as the Split Depth-wise Transpose Attention (STDA) encoder. This encoder works by dividing input tensors into several channel groups. It then uses depth-wise convolution in conjunction with self-attention mechanisms across the channel dimensions. By doing so, the STDA encoder naturally enlarges the receptive field and effectively encodes features at multiple scales. The extensive requirements of the transformer self-attention layer make it impractical for vision tasks on edge devices, primarily due to its high MAdds and latency. To address this issue in SDTA encoder, they utilize transposed query and key attention feature maps \cite{ali2021xcit}. This approach enables linear complexity by performing the dot-product operation of the Multi-Head Self-Attention (MSA) across channel dimensions, instead of spatial dimensions. As a result, cross-covariance across channels can be computed and create attention feature maps that inherently contain global representations. They also introduce adaptive kernel sizes to capture more global information by using smaller kernel sizes in the initial layers followed by larger kernels for the latter stages in the convolutional encoder stages.  These models come in various sizes, offering flexibility based on specific requirements. They include the extra-extra small, extra-small, and small variants. More details about the architecture can be found in \cite{maaz2023edgenext}. Details of the variants and dimensions of feature maps at different levels are shown in Table \ref{tab:layers}.

\subsection{Low Rank Linear Module (\emph{LoRaLin})}
Despite the considerable optimization offered by the EdgeNeXt architecture, it is observed that a significant portion of both computational and parameter overhead originates from the linear layers. In an attempt to attenuate these parameter demands, we propose the incorporation of a Low Rank Linear Module (\emph{LoRaLin}). This module effectively reduces computational requirements while maintaining minimal compromise to overall performance.

Hu \textit{et al.} \cite{hu2021lora} proposed an approach termed Low-Rank Adaptation (LoRA) for reducing the number of trainable parameters in large language models during fine-tuning. The LoRA tuning method maintains the weights of the pretrained model unchanged while introducing trainable rank decomposition matrices into every layer of the Transformer architecture. This technique draws inspiration from the concept of `low intrinsic dimension` observed when adapting a pretrained model to a specific task \cite{aghajanyan2020intrinsic}. The amount of newly introduced parameters are considerably less, even though the original full rank matrices needs to be used at inference time. However, our aim is to reduce the parameter count of the model while accepting a trade-off in terms of model capacity. To accomplish this, we adopt a strategy of factorizing each fully connected layer into two low rank matrices.

Consider a fully connected layer in the network:

\begin{equation}
   Y = W_{M \times N}X + b.
\end{equation}

The weight matrix \( W \) in a linear layer of a neural network, which maps an input of size \( M \) to an output of size \( N \), has dimensions \( M \times N \).

This matrix can be represented as the product of two low rank matrices as follows:

\begin{equation}
   W_{M \times N} = W_{M \times r} \cdot W_{r \times N},
\end{equation}
where, $W_{M \times r}$ and $W_{r \times N}$, are low rank matrices with a rank $r$. 

Now, the original linear layer can be implemented as :

\begin{equation}
   Y = W_{r \times N}( W_{M \times r} (X))+b.
\end{equation}
Essentially as two linear layers with lower ranks, this reduces the number of parameters, and the number of multiply adds (MAdds). 

This can be implemented using two linear layers instead of one as shown in Fig. \ref{list:listing1}.
\begin{figure}[htb!]
\caption{PyTorch class for a Low-Rank Linear layer (\emph{LoRaLin}).}
\begin{lstlisting}[language=Python]
class LoRaLin(nn.Module):
    def __init__(self, in_feat, out_feat, gamma, bias):
        super(LoRaLin, self).__init__()
        rank = max(2,min(in_feat, out_feat) * gamma))
        self.lin1 = nn.Linear(in_feat, rank, bias=False)
        self.lin2 = nn.Linear(rank, out_feat, bias=bias)

    def forward(self, input):
        x = self.lin1(input)
        x = self.lin2(x)
        return x
\end{lstlisting}
\label{list:listing1}
\end{figure}

In this context, the rank of each module is determined by a hyper parameter known as Rank-ratio ($\gamma$), which governs the ratio between the ranks. A minimum value of two is employed as the lower limit for the rank in our implementation.

\begin{equation}
   rank= \max(2, \gamma * \min(M, N)),
\end{equation}

\begin{figure}[h]
    \centering
    \includegraphics[width=0.45\textwidth]{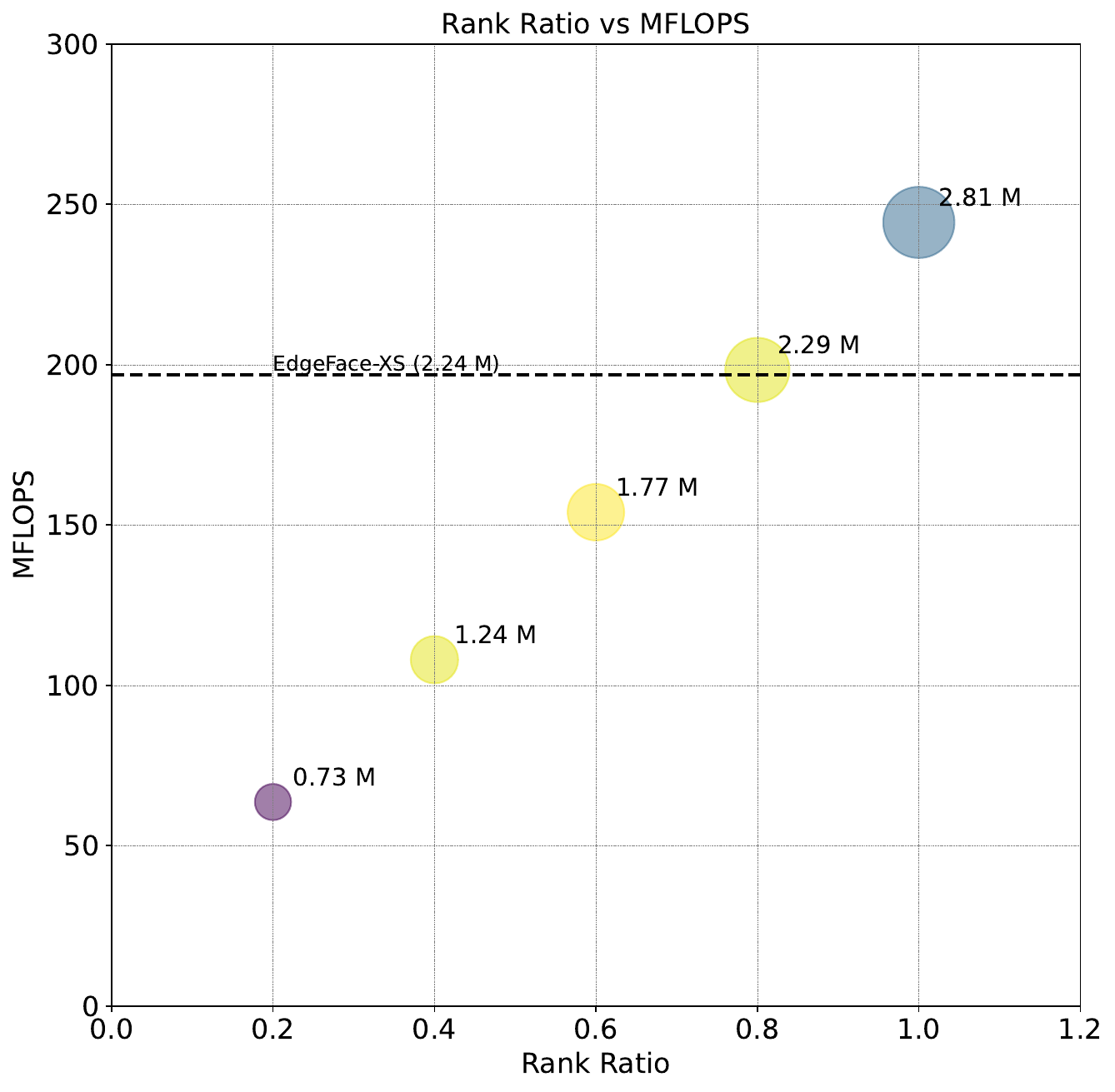} 
    \caption{The figure shows the reduction in Model Parameters (MPARAMS) and Multiply-Accumulate Operations (MFLOPS) as a function of Rank-ratio ($\gamma$). The dotted line represents the corresponding values from the default model employing a conventional Linear layer.}
    \label{fig:rankplot}
\end{figure}

By varying the value of $\gamma$, both the number of parameters and FLOPS undergo changes. For instance, in the case of the ``edgenext-extra-small (XS)'' network, the Figure \ref{fig:rankplot} illustrates the reduction in the number of parameters and FLOPS with lower values of $\gamma$. The dotted line represents the values associated with the original linear layer. Notably, for $\gamma \le 0.8$, both parameter count and computational efficiency demonstrate improvements compared to the base model.

\subsection{Training details}

The dataset used for training the FR models constitutes selected subsets of the Webface260M dataset \cite{zhu2021webface260m}, specifically, the WebFace 12M and WebFace 4M subsets. These subsets are characterized by an abundance of pre-aligned face images, each with a resolution of $112 \times 112$. The initial preprocessing step entails the conversion of these images into tensors, followed by normalization within the -1 to 1 range. We further enhance the data variability through a series of augmentations, including random grayscale conversion, resizing, and blurring. These augmentations are implemented leveraging the capabilities of the DALI \cite{dali} library. The models were trained with 4/8  Nvidia RTX 3090 (24GB) GPUs using distributed training strategy. We trained our models using PyTorch  with AdamW optimizer \cite{loshchilov2017decoupled} and trained the models with CosFace \cite{wang2018cosface} loss function using a polynomial decay learning rate schedule with restarts to achieve the best performance. The batchsize on a single GPU varied from 256 to 512 depending on the size of the model.  The embedding size during training is kept as 512. We used the distributed PartialFC algorithm \cite{an2022killing} for faster training and to handle memory issues while dealing with a large number of identities. During inference, the classification head is removed and the resulting 512-D embedding is used for the comparisons. The training settings and hyper parameters for different models were selected for optimal performance.

\section{Experiments}\label{sec:experiment}

\subsection{Test Datasets}
We evaluated performance of the proposed \model model on seven distinct benchmarking datasets. The datasets selected for assessment include Labeled Faces in the Wild (LFW) \cite{huang2008labeled}, Cross-age LFW (CA-LFW) \cite{zheng2017cross}, CrossPose LFW (CP-LFW) \cite{zheng2018cross}, Celebrities in Frontal-Profile in the Wild (CFP-FP) \cite{sengupta2016frontal}, AgeDB-30 \cite{moschoglou2017agedb}, IARPA Janus Benchmark-B (IJB-B) \cite{whitelam2017iarpa}, and IARPA Janus Benchmark-C (IJB-C) \cite{maze2018iarpa}.
To maintain consistency with prior works, we report accuracy values for high-resolution datasets such as LFW, CA-LFW, CP-LFW, CFP-FP, and AgeDB-30.
For the IJB-B and IJB-C datasets, we report the True Accept Rate (TAR) at a False Accept Rate (FAR) of 1e-4.

\input{perf_table}

\begin{figure}[tbhp]
\centering
\begin{subfigure}[b]{0.875\linewidth}
\centering
    \includegraphics[width=0.99\columnwidth]{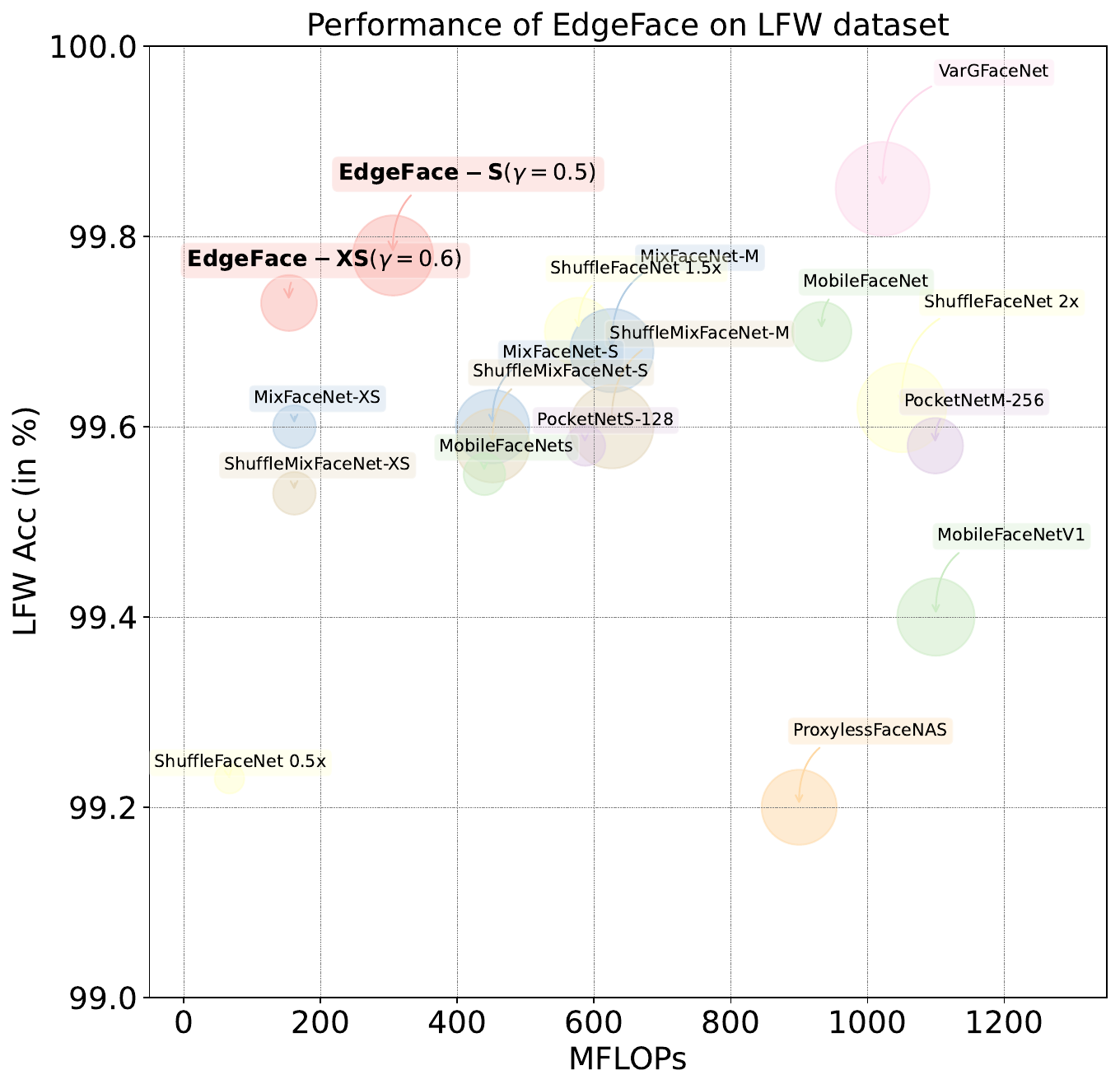} 
\caption{LFW}
\end{subfigure}
\\
\begin{subfigure}[b]{0.875\linewidth}
\centering
    \includegraphics[width=0.99\columnwidth]{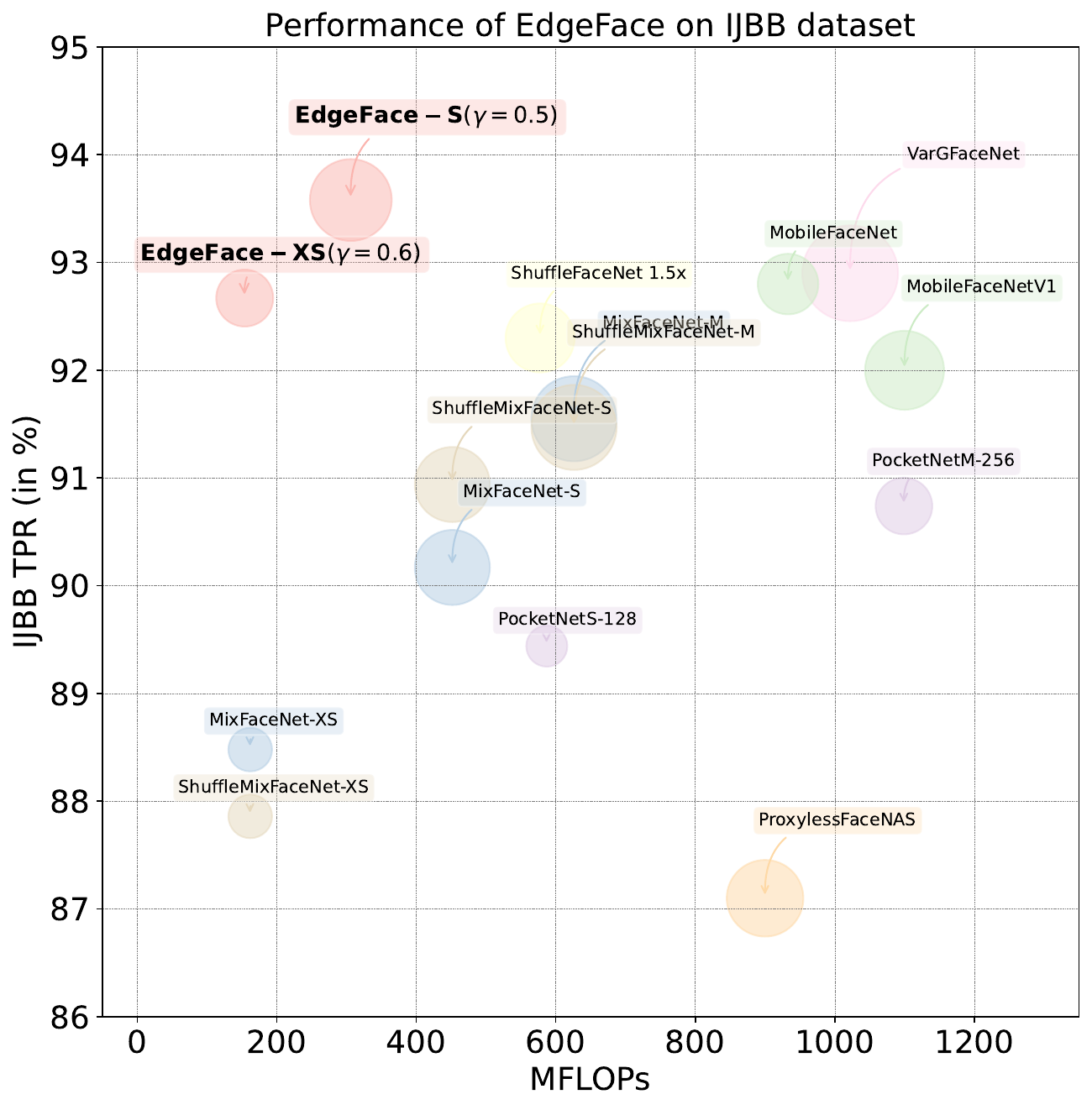} 
\caption{IJB-B}
\end{subfigure}
\\
\begin{subfigure}[b]{0.875\linewidth}
\centering
    \includegraphics[width=0.99\columnwidth]{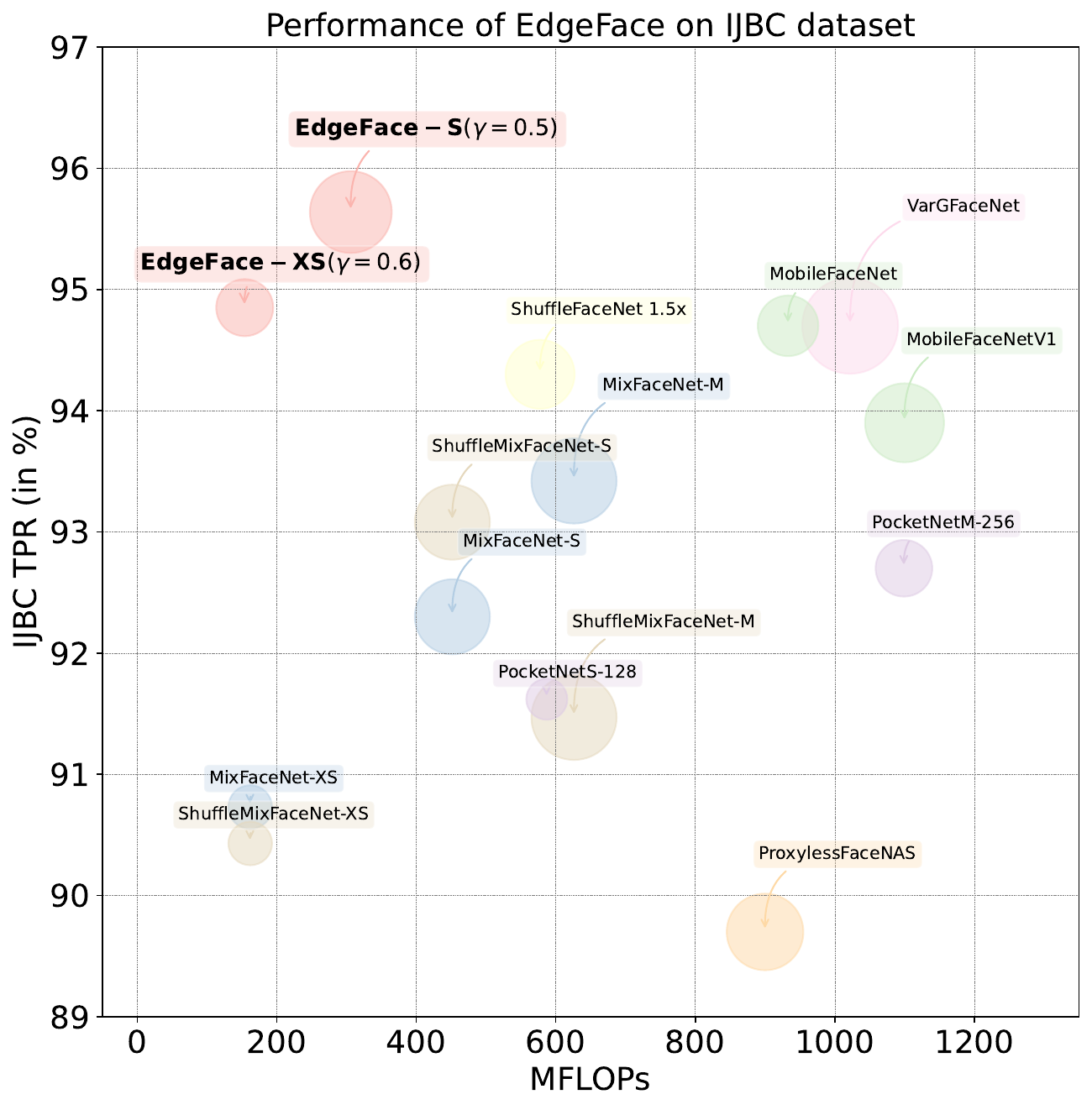} 
\caption{IJB-C}
\end{subfigure}
\vspace{-5pt}
\caption{Performance comparison of different models on (a) LFW, (b) IJB-B, and (c) IJB-C datasets.}
\label{fig:perf_comparison}
\end{figure}

\subsection{Comparison with SOTA} \label{subsec:exp:comparison}
Table~\ref{tab:verification} compares  our method with SOTA lightweight face recognition models in the literature on different benchmarking datasets. We categorized models in the literature based on the number of parameters into 2-5 M parameters  and $<\text{2M}$ parameters. 
In each category, we also have a representative version of \model model.
In the category of 2-5 M parameters models, our representative model is EdgeFace-S ($\gamma = 0.5$), and in the second category ($<\text{2M}$ parameters) we can consider EdgeFace-XS ($\gamma = 0.6$) as our representative model.
As the results in this table show our \model models achieve competitive performance with SOTA lightweight models in the literature.
For CA-LFW, CP-LFW, IJB-B, and IJB-C datasets, our EdgeFace-S ($\gamma = 0.5$) model achieves the best recognition accuracy compared to SOTA models 2-5 M parameters. It is noteworthy our EdgeFace-S ($\gamma = 0.5$) model is also the most efficient model in terms of FLOPs among the SOTA lightweight models with  2-5 M parameters.
For the second category, our EdgeFace-XS ($\gamma = 0.6$) model achieves the best recognition performance for LFW, CP-LFW, IJB-B, and IJB-C datasets. Compared to other models in the same category, our model is the second most efficient model in terms of FLOPs. In this category, we observe that ShuffleFaceNet 0.5x has fewer FLOPs, but it also has the poorest recognition performance in all datasets. The superior performance of our models in terms of FLOPS to performance can be observed in Fig.~\ref{fig:perf_comparison}.

\subsection{Ablation studies }

\subsubsection{Ablation with varying values of $\gamma$}

To evaluate the effectiveness of the \textit{LoRaLin} layers, we conducted a series of experiments using the EdgeFace-XS model. These experiments involved varying the value of $\gamma$ from 0.2 to 1, with increments of 0.2. All models were trained using the same configuration for 50 epochs. As a point of reference, we also compared these models with the \textit{default} EdgeFace-XS model, which does not include the \textit{LoRaLin} layer.

Figure \ref{fig:rankplot} illustrates the changes in model parameters and FLOPs as the value of $\gamma$ varies. It is observed that the parameters and FLOPs remain consistent with the EdgeFace-XS model when $\gamma$ is approximately 0.8. For values of $\gamma$ below 0.8, there is a reduction in model parameters, FLOPs, and size. 

\begin{figure}[h]
\centering
\includegraphics[width=0.95\columnwidth]{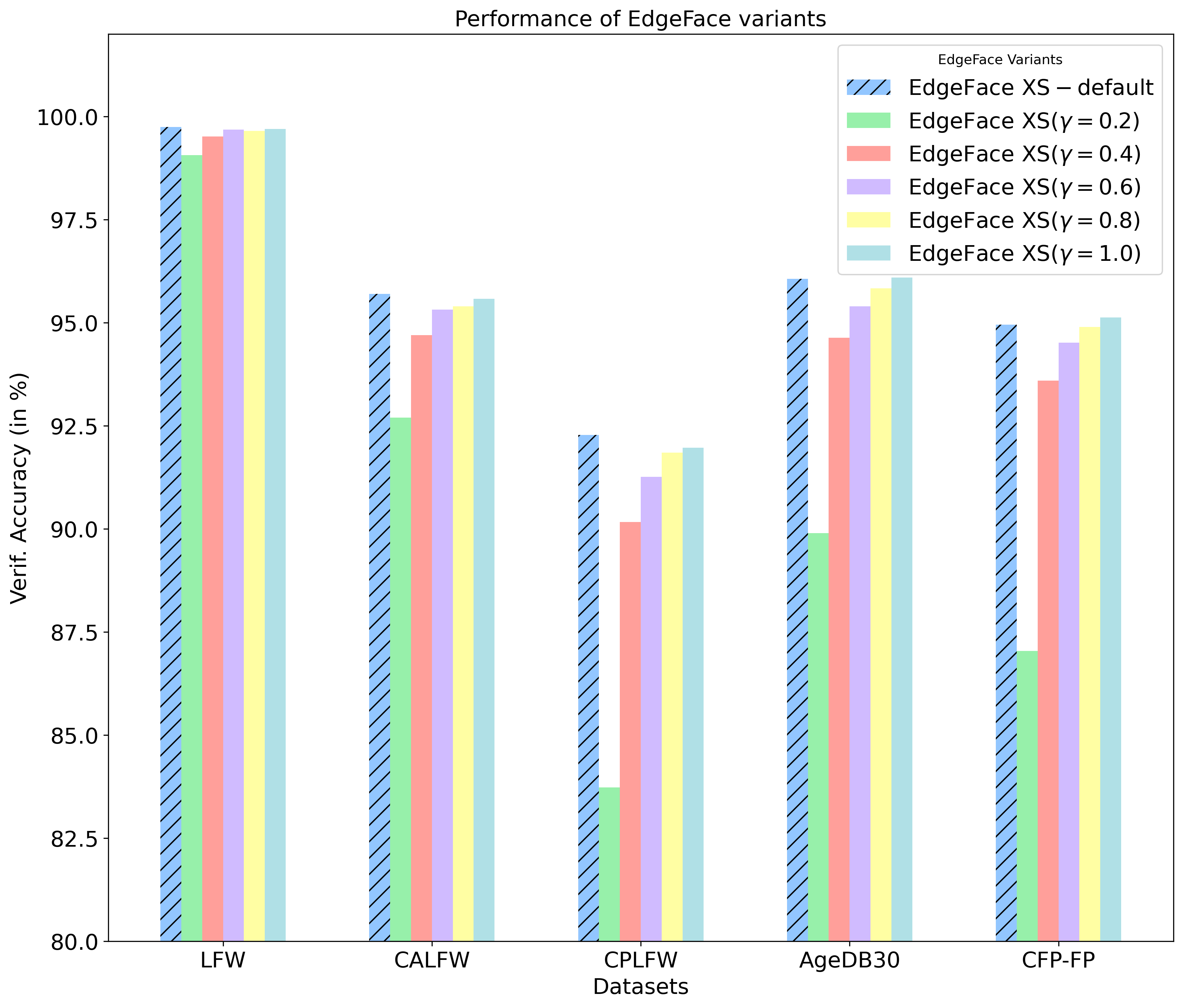} 
\caption{Ablation study of \model with respect to low rank parameter ($\gamma$). The performance is evaluated as the verification accuracy of `XS' variant on different face recognition datasets.}
\label{fig:edgeface_ablations}
\end{figure}

To assess the performance of these models, we evaluated them using standard benchmarks. The results are presented in Table \ref{tab:ablation_perf}, which displays the performance across these benchmarks. The performance deteriorates as the value of $\gamma$ decreases (Fig. \ref{fig:edgeface_ablations}). However, it is notable that the performance remains satisfactory up to $\gamma = 0.6$, beyond which it starts to decline more sharply.

\begin{table}
	\setlength{\tabcolsep}{2.35pt}
\caption{The comparison of performance of the default and low rank $\gamma=0.6$ variants of EdgeFace-XS. The $\%$ difference in verification accuracy as well as in parameters and FLOPS are provided in the table.}
\resizebox{\columnwidth}{!}{

\begin{tabular}{l | r | r | r | r | r }
\toprule

\textbf{Model} & LFW & IJB-B & IJB-C & \textbf{MPARAMS} & \textbf{MFLOPS} \\
\midrule
EdgeFace-XS & 99.8 & 92.65 & 94.75 & 2.24 & 196.9 \\ 

$\gamma=0.6$ & 99.7 (0.1\% $\downarrow$) & 92.24 (\textbf{0.4}\%$\downarrow$)) & 94.28 (\textbf{0.49}\%$\downarrow$)) & \textbf{1.77} (\textbf{21}\% $\downarrow$) & \textbf{153.9} (\textbf{22}\% $\downarrow$) \\
\bottomrule
\end{tabular}
}

\label{tab:ablation_improvement}
\end{table}

\begin{figure}[htb!]
    \centering
    \includegraphics[width=\columnwidth]{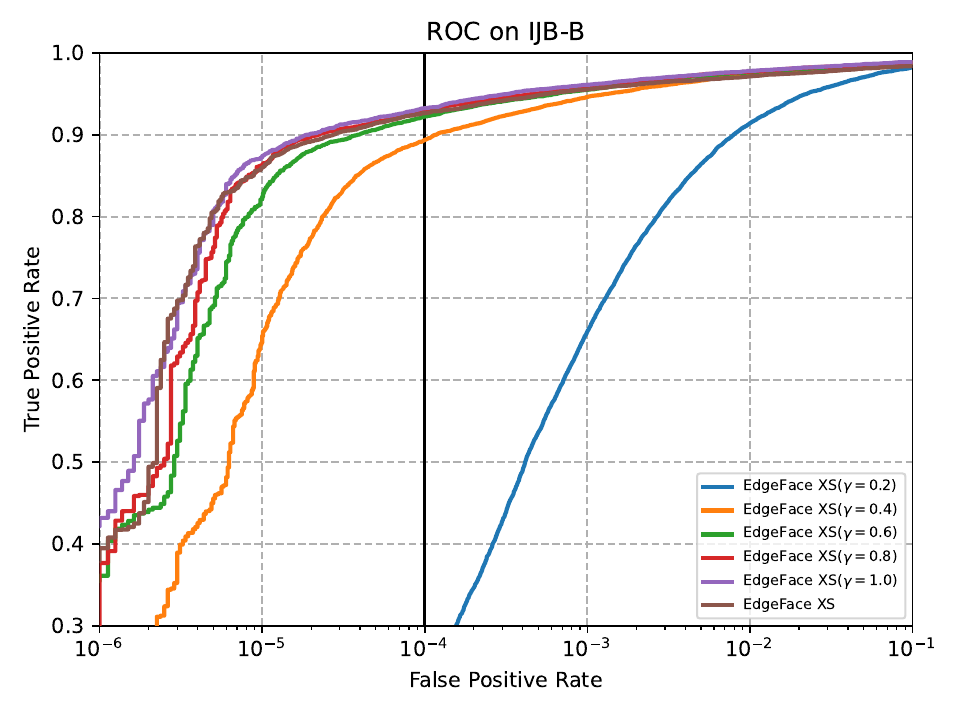}\\%
    \includegraphics[width=\columnwidth]{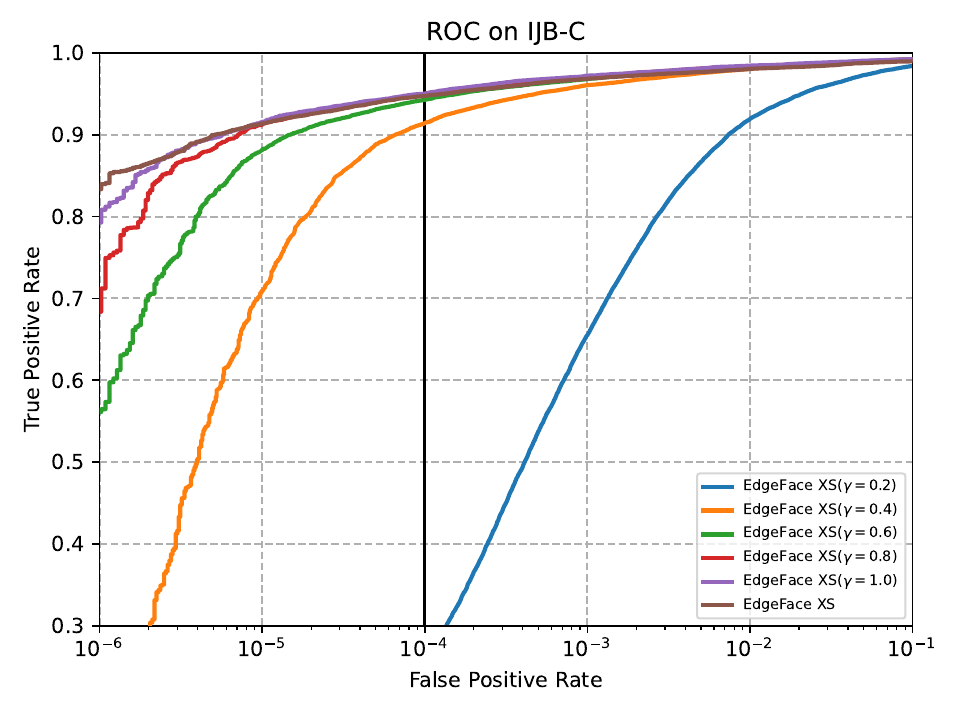}
    \caption{Performance evolution for EdgeFace-XS across different Rank-ratio $\gamma$ on IJB-B and IJB-C datasets.}
    \label{fig:edgeface_ijb_ablation}
\end{figure}

Figure \ref{fig:edgeface_ijb_ablation} demonstrates the performance changes of the models on the IJB-B and IJB-C datasets. In both cases, the proposed method achieves good performance up to $\gamma = 0.6$. Additionally, Table \ref{tab:ablation_improvement} provides the percentage points of performance degradation corresponding to the changes in model parameters and FLOPs for the IJB-C and IJB-B datasets. It can be seen that we can obtain around 20\% savings in parameters and FLOPS with less than 0.5\% drop in accuracy.

The results presented in Table \ref{tab:ablation_improvement} highlight that our approach achieves a significant improvement in parameter and FLOP efficiency while maintaining a minimal reduction in performance. This demonstrates the effectiveness of our approach in achieving a favorable trade-off between efficiency and performance.

\input{ablation_table}

\subsubsection{Ablation studies with loss functions}

In this section, we perform experiments with training the \model with different loss functions; specifically, we train the same network with ArcFace \cite{deng2019arcface} and CosFace \cite{wang2018cosface} loss functions. We use the XS variant of the \model with $\gamma=0.6$ as selected from the previous section for this set of experiments. We retrained the same model with ArcFace and CosFace for 50 epochs with a batch size of 512. All the hyperparameters were kept the same for both cases for a fair comparison. The results from these experiments are shown in Table \ref{tab:ablation_loss}. In high-resolution benchmarks, ArcFace performs better on LFW, CP-LFW, and CFP-FP, while CosFace performs better on CA-LFW and AgeDB-30. In the evaluations with IJB-B and IJB-C, it can be seen that CosFace slightly performs better than ArcFace at an FPR of $1e-4$, but ArcFace performs much better than CosFace on very low FPR regions ($1e-6$).

\begin{table}
\centering
\caption{The performance of ablation of \textbf{XS} variant of \model on different face recognition datasets. The ablation is performance with respect to loss function used in the training. For top 5 rows, all values indicate verification accuracy (TAR) expressed as percentages; while for IJB-B and IJB-C datasets, the values reported are true positive rate (TPR) at false positive rate (FPR) of $1e-4$ and $1e-6$.}
{

\resizebox{0.70\columnwidth}{!}{%
\begin{tabular}{l | r | r }
\toprule
\textbf{Dataset} & ArcFace \cite{deng2019arcface} & CosFace \cite{wang2018cosface} \\
\midrule
LFW     &\textbf{99.65}   &99.53 \\
CA-LFW  &95.07   &\textbf{95.38} \\
CP-LFW  &\textbf{91.50}   &91.35 \\
AgeDB-30&95.58   &\textbf{95.60} \\
CFP-FP  &\textbf{94.80}   &94.26 \\ \midrule
IJB-B [FAR@ $1e-4$]   &91.69   &\textbf{92.41} \\
IJB-B [FAR@ $1e-6$]   &\textbf{38.94} &30.20 \\ \midrule
IJB-C [FAR@ $1e-4$]   &94.09   &\textbf{94.39} \\ 
IJB-C [FAR@ $1e-6$]   &\textbf{75.47}   &55.46   \\

\bottomrule
\end{tabular}
}
}
\label{tab:ablation_loss}
\end{table}

\subsection{IJCB'23 Efficient Face Recognition (EFaR) competition }
\label{subsec:exp:IJCB}
The 2023 International Joint Conference on Biometrics (IJCB 2023) featured the Efficient Face Recognition Competition (EFaR), which aimed to stimulate advancements in efficient face recognition techniques. Rankings were determined using a composite score that factored in verification accuracy across a variety of benchmarks, as well as model deployability metrics such as floating-point operations and model size. The contest consists of two categories, differentiated by the model's parameter count. Teams have the flexibility to present two solutions per category. First category is dedicated to highly compact networks possessing fewer than 2 million parameters (denoted as $ < $2 MP). The latter category addresses models that have parameters ranging between 2 and 5 million (2--5 MP). Entries are assessed and ranked within their respective categories. For both categories, a submission's rank is determined by its verification accuracy, computational intricacy, and memory footprint of the model.

For evaluation of face recognition performance the competition organizers utilized accuracy on LFW, CPLFW, CALFW, CFP-FP, and AgeDB30 datasets. Additionally, they also used IJB-C dataset, where they applied the true acceptance rate (TAR) at a false acceptance rate (FAR) of \(10^{-4}\), denoted as TAR at FAR=\(10^{-4}\). A cumulative ranking across all benchmarks is calculated using the aggregated Borda counts from each dataset, which is then used as the final ranking.
To assess a solution's deployability, the competition organizers factor in the model's compactness (measured by the number of parameters), its memory usage (indicated by the model size in MB), and its computational demand (denoted by M FLOPs). For all these aspects, a lower figure suggests superior deployability. Teams were required to provide these metrics (number of parameters, model size, FLOPs) for their entries. Rankings are formulated based on FLOPs and model dimensions. The final team ranking for a track is based on a weighted Borda count incorporating: (a) the standardized Borda count from the evaluated benchmarks, (b) the Borda count associated with the FLOPs metrics, and (c) the Borda count for the model size. The performance on benchmarks holds a weightage of 70\%, while both the FLOPs and the model size each carry a weightage of 15\%.

Teams had the liberty to present two entries for both tracks. Altogether, teams contributed 17 unique submissions. Out of these, 9 solutions were for the 2-5 M parameter category, while the $<2 $ M parameter category received 8 solutions.

\input{competition_perf}

Two \model variants were submitted to each of the categories, they are:

\textbf{For 2-5M parameters:}
\begin{enumerate}
    \item Idiap EdgeFace-S ($\gamma$=0.5) : This is the `small` variant of \model model with $\gamma$=0.5 in the LoRaLin layers. The parameters are float32 precision.
    \item Idiap EdgeFace-XS-Q :  This is the `extra-small` variant of \model model without LoRaLin layers. The linear layers of the model are quantized to 8-bit in this case.
\end{enumerate}

\textbf{For less than 2M parameters:}
\begin{enumerate}
    \item Idiap EdgeFace-XS ($\gamma$=0.6):  This is the `extra-small` variant of \model model with $\gamma$=0.6 in the LoRaLin layers. The parameters are float32 precision.
    \item Idiap EdgeFace-XXS-Q: This is the `extra-extra-small` variant of \model model without LoRaLin layers. The linear layers of the model are quantized to 8-bit in this case.
\end{enumerate}

The results from the competition are tabulated in Table \ref{tab:competition_table} (reproduced from the competition paper \cite{kolf2023efar}). For the submitted solutions with $<2$ million parameters,  Idiap EdgeFace-XS ($\gamma$=0.6) \textbf{is the overall best performing model}. Idiap EdgeFace-XS($\gamma$=0.6) also ranks first when considering the verification accuracy, including first rank on IJB-C. From the results of the experiment of the models with 2-5 million parameters, although it does not ranked first, highest ranked solution in terms of verification accuracy over the evaluated benchmarks is Idiap EdgeFace-S ($\gamma$=0.5). It also achieved the highest TAR at FAR=$10^{-4}$ on the large IJB-C benchmark for all submitted solutions in this category. It can be seen that variants of our \model achieves superior performance in terms of other models in both categories. Also, for the  highly compact networks possessing fewer than 2 million parameters, our model ranks first in the competition, showing the effectiveness of our approach. More details of the competition and evaluation results can be found in the competition paper \cite{kolf2023efar}.

\section{Discussions }
\label{sec:discussions}
Our experiments in Section~\ref{subsec:exp:comparison} show that our model is very efficient and also achieves competitive recognition  accuracy compared to SOTA lightweight models. 
Among seven benchmarking datasets used in our evaluation, \model achieves the best recognition performance for four different datasets in each of the categories of models with 2-5 M parameters  and $<\text{2M}$ parameters. Achieving such a high recognition accuracy is more particularly impressive considering the computation of different models in terms of FLOPs in Table~\ref{tab:verification}, where we observe that \model is the most efficient model in the first category (2-5 M parameters) and the second most efficient model in the second category ($<\text{2M}$ parameters).

Among our different benchmarking datasets, five datasets (i.e., LFW, CA-LFW, CP-LFW, CFP-LFW, and AgeDB-30) have higher-quality face images. The results in Table~\ref{tab:verification} show that our model achieves competitive performance with SOTA models on these benchmarking datasets.
In contrast,
IARPA Janus Benchmark datasets (i.e., IJB-B and IJB-C) include images with different qualities  (including low-quality images)  and are among the most challenging face recognition benchmarking datasets. According to the results in Table~\ref{tab:verification}, \model outperforms all previous lightweight models in both categories of models with 2-5 M parameters  and $<\text{2M}$ parameters on these two datasets, which shows the superiority of our model for different quality of images.

Last but not least,  we would like to highlight that, as mentioned in Section~\ref{subsec:exp:IJCB}, variations of \model achieved best verification accuracy in both categories of $<2$M parameters and 2-5M parameters in the recent efficient face recognition competition in IJCB 2023~\cite{kolf2023efar} amongst all submissions, which used state-of-the-art techniques to train efficient face recognition models. In particular, for the category of models with less than two million parameters, our model not only was the first model in terms of verification accuracy, but also received the first place overall in terms of  verification accuracy, computation complexity and memory footprint.


\section{Conclusions}
\label{sec:conclusions}

In this paper, we introduced \model\!,  a highly efficient face recognition model that combines the strengths of CNN and Transformers. By leveraging efficient hybrid architecture and \textit{LoRaLin} layers, the \model model achieves remarkable performance while maintaining low computational complexity. Our extensive experimental evaluations on various face recognition benchmarks, including LFW, AgeDB-30, CFP-FP,  IJB-B, and IJB-C, demonstrate the effectiveness of \model\!. Our hybrid design strategy incorporates convolution and efficient self-attention-based encoders, providing an ideal balance between local and global information processing. This enables \model to achieve superior performance compared to state-of-the-art methods while maintaining low parameters and MAdds. The proposed \model model secured the first position in general ranking among models having less than 2M parameters in the IJCB 2023 Efficient Face Recognition Competition \cite{kolf2023efar}. In addition, in both categories of models with 2--5 M and $<2$ M parameters, variants of \model achieved the best verification accuracy in the IJCB 2023 competition. In summary, \model\! offers an efficient and highly accurate face recognition model tailored for edge devices. Knowledge distillation strategies can further enhance the model's performance, while exploring different quantization methods holds potential for improving storage and inference, which can be pursued in future research.

\section*{Acknowledgements}
This research is partly based upon work supported by the H2020 TReSPAsS-ETN Marie Sk\l{}odowska-Curie early training network (grant agreement 860813), as well as based on the work supported by the Hasler foundation through the SAFER project.
 \\

This research is also based upon work supported in part by the Office of the Director of National Intelligence (ODNI), Intelligence Advanced Research Projects Activity (IARPA), via [2022-21102100007]. The views and conclusions contained herein are those of the authors and should not be interpreted as necessarily representing the official policies, either expressed or implied, of ODNI, IARPA, or the U.S. Government. The U.S. Government is authorized to reproduce and distribute reprints for governmental purposes notwithstanding
 any copyright annotation therein.

\bibliographystyle{IEEEtran}

\bibliography{sn-bibliography}

\begin{IEEEbiography}[{\includegraphics[width=1in,height=1.25in, trim={0cm 0.5cm 0cm 0.3cm},clip,keepaspectratio]{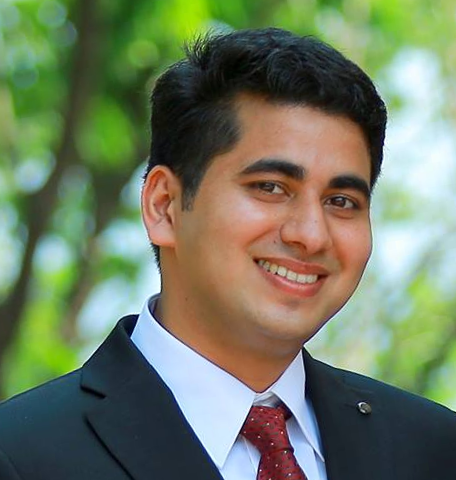}}]{Anjith George} has received his Ph.D. and M-Tech degree from the Department of Electrical Engineering, Indian Institute of Technology (IIT) Kharagpur, India in 2012 and 2018 respectively. After Ph.D, he worked in Samsung Research Institute as a machine learning researcher. Currently, he is a research associate in the biometric security and privacy group at Idiap Research Institute, focusing on developing face recognition and presentation attack detection algorithms. His research interests are real-time signal and image processing, embedded systems, computer vision, machine learning with a special focus on Biometrics.
\end{IEEEbiography}
\vspace{-10mm}
\begin{IEEEbiography}[{\includegraphics[width=1in,height=1.25in, trim={0cm 0.15cm 0cm 0cm},clip,keepaspectratio]{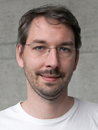}}]{Christophe Ecabert}
 received his Ph.D. and M.Sc degrees in electrical engineering from the \'{E}cole Polytechnique F\'ed\'erale de Lausanne (EPFL) in 2021 and 2014 respectively. He is currently a Post-Doctoral Assistant with the Biometrics Security and Privacy Group at Idiap Research Institute. His current work aims at exploring the use of synthetic data in the context of face recognition.
\end{IEEEbiography}
\vspace{-10mm}
\begin{IEEEbiography}[{\includegraphics[width=1in,height=1.25in, trim={0cm 0.5cm 0cm 0.3cm},clip,keepaspectratio]{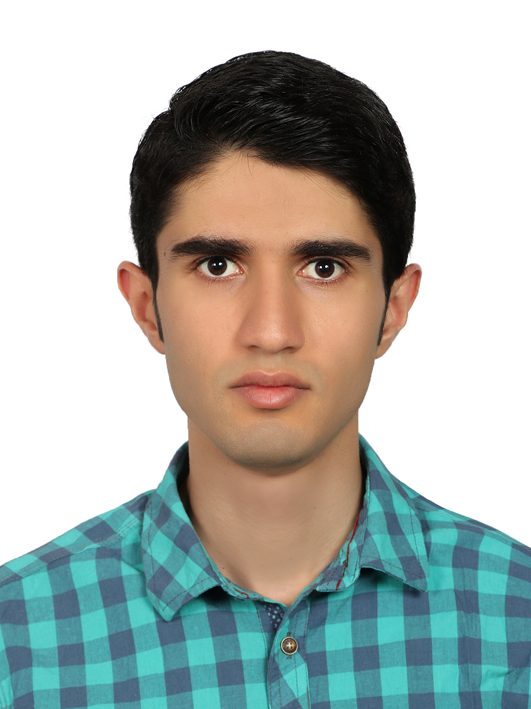}}]{Hatef Otroshi Shahreza} 
	received the B.Sc. degree (Hons.) in electrical engineering from the University of Kashan, Iran, in 2016, and the M.Sc. degree in electrical engineering from the Sharif University of Technology, Iran, in 2018. He is currently pursuing the Ph.D. degree with the \'{E}cole Polytechnique F\'{e}d\'{e}rale de Lausanne (EPFL), Switzerland, and is a Research Assistant with the Biometrics Security and Privacy Group, Idiap Research Institute, Switzerland, where he received H2020 Marie Sk\l{}odowska-Curie Fellowship (TReSPAsS-ETN) for his doctoral program. During his Ph.D., Hatef also experienced 6 months as  a visiting scholar with the Biometrics and Internet Security Research Group at Hochschule Darmstadt, Germany.  He is also the winner of the European Association for Biometrics (EAB) Research Award 2023. 
	His research interests include deep learning, computer vision, and biometrics.
\end{IEEEbiography}
\vspace{-10mm}
\begin{IEEEbiography}[{\includegraphics[width=1in,height=1.25in, trim={0cm 0.5cm 0cm 0.3cm},clip,keepaspectratio]{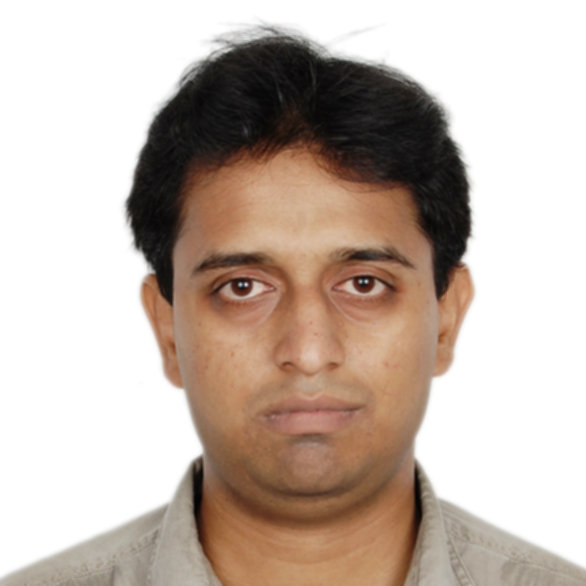}}]{Ketan Kotwal}
received the M.Tech and Ph.D. degrees in electrical engineering from the Indian Institute of Technology Bombay (IIT Bombay, Mumbai, India).
His current research interests include various topics in
image processing, machine learning, and data processing. He has also
been actively involved in technology consulting in relevant areas.
He received the Excellence in Ph.D. Thesis Award from the IIT
Bombay, and the Best Ph.D. Thesis Award from the Computer Society
of India for his doctoral work. Dr. Kotwal is a co-author of research
monograph ``Hyperspectral Image Fusion'' (Springer, US). At present,
he is a member of the Biometrics Security and Privacy Group at the
Idiap Research Institute.
\end{IEEEbiography}
\vspace{-10mm}
\begin{IEEEbiography}[{\includegraphics[width=1in,height=1.25in,trim={7cm 0cm 7cm 0.5cm},clip,keepaspectratio]{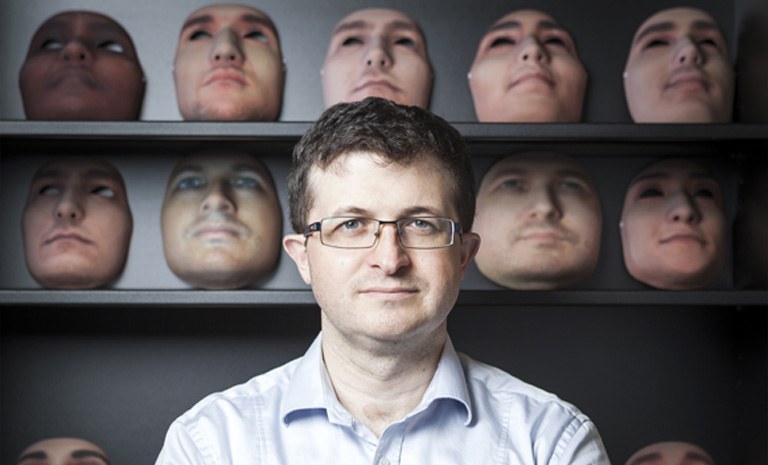}}]{S{\'e}bastien Marcel} heads the Biometrics Security and Privacy group at Idiap Research Institute (Switzerland) and conducts research on face recognition, speaker recognition, vein recognition, attack detection (presentation attacks, morphing attacks, deepfakes) and template protection. He received his Ph.D. degree in signal processing from Universit{\'e} de Rennes I in France (2000) at CNET, the research center of France Telecom (now Orange Labs). He is Professor at the University of Lausanne (School of Criminal Justice) and a lecturer at the  \'{E}cole Polytechnique F{\'e}d{\'e}rale de Lausanne. He is also the Director of the Swiss Center for Biometrics Research and Testing, which conducts certifications of biometric products.
\end{IEEEbiography}





\end{document}

%% file: perf_table.tex
\begin{table*}[h]
\centering
\caption{Performance evaluation (TAR) of the proposed EdgeFace model, along with various recent compact FR models, on 7 benchmarking datasets. 
The models are ordered based on the number of parameters. All decimal points are provided as reported in the respective works. Models are categorized based on the number of parameters into 2-5M  parameters and $<\text{2M}$ parameters. For each benchmarking dataset, the best performance in each category is emboldened.}
\resizebox{\linewidth}{!}{%
\begin{tabular}{l|c|c|c|c|c|c|c|c|cc} 

\toprule
\multirow{2}{*}{\textbf{Model}} & \multirow{2}{*}{\textbf{MPARAMS}} & \multirow{2}{*}{\textbf{MFLOPs}} & \multirow{2}{*}{\begin{tabular}[c]{@{}c@{}}\textbf{LFW} \\ (\textbf{\%})\end{tabular}} 
& \multirow{2}{*}{\begin{tabular}[c]{@{}c@{}}\textbf{CA-LFW}\\ (\textbf{\%})\end{tabular}} & \multirow{2}{*}{\begin{tabular}[c]{@{}c@{}}\textbf{CP-LFW}\\ (\textbf{\%})\end{tabular}} & \multirow{2}{*}{\begin{tabular}[c]{@{}c@{}}\textbf{CFP-FP}\\ (\textbf{\%})\end{tabular}} & \multirow{2}{*}{\begin{tabular}[c]{@{}c@{}}\textbf{AgeDB-30}\\ (\textbf{\%})\end{tabular}} & \multirow{2}{*}{\begin{tabular}[c]{@{}c@{}}\textbf{IJB-B}\\ (\textbf{\%})\end{tabular}} & \multirow{2}{*}{\begin{tabular}[c]{@{}c@{}}\textbf{IJB-C}\\ (\textbf{\%})\end{tabular}}  \\ 
                       &                             &                         &                                                                      &                         &                                                                        &                                                                        &                                                                          &                                                                       &                                                                      \\ 
\midrule
VarGFaceNet  \cite{yan2019vargfacenet,martinez2021benchmarking}          & 5.0                         & 1022                    & \textbf{99.85}                                                                & 95.15                   & 88.55                                                                  & \textbf{98.50}                                                                  & \textbf{98.15}                                                                    & 92.9                                                                  & 94.7                  \\ 
ShuffleFaceNet 2×  \cite{martindez2019shufflefacenet}     & 4.5                         & 1050                    & 99.62                                                                & -                       & -                                                                      & 97.56                                                                  & 97.28                                                                    & -                                                                     & -     \\ 
MixFaceNet-M \cite{boutros2021mixfacenets}          & 3.95                        & 626.1                   & 99.68                                                                & -                       & -                                                                      & -                                                                      & 97.05                                                                    & 91.55                                                                 & 93.42     \\ 
ShuffleMixFaceNet-M  \cite{boutros2021mixfacenets}   & 3.95                        & 626.1                   & 99.60                                                                & -                       & -                                                                      & -                                                                      & 96.98                                                                    & 91.47                                                                 & 91.47     \\ 
MobileFaceNetV1  \cite{martinez2021benchmarking}      & 3.4                         & 1100                    & 99.4                                                                 & 94.47                   & 87.17                                                                  & 95.8                                                                   & 96.4                                                                     & 92.0                                                                  & 93.9    \\ 
ProxylessFaceNAS  \cite{martinez2021benchmarking}     & 3.2                         & 900                     & 99.2                                                                 & 92.55                   & 84.17                                                                  & 94.7                                                                   & 94.4                                                                     & 87.1                                                                  & 89.7  \\ 
MixFaceNet-S    \cite{boutros2021mixfacenets}        & 3.07                        & 451.7                   & 99.6                                                                 & -                       & -                                                                      & -                                                                      & 96.63                                                                    & 90.17                                                                 & 92.30     \\ 
ShuffleMixFaceNet-S \cite{boutros2021mixfacenets}    & 3.07                        & 451.7                   & 99.58                                                                & -                       & -                                                                      & -                                                                      & 97.05                                                                    & 90.94                                                                 & 93.08      \\ 
ShuffleFaceNet 1.5x  \cite{martindez2019shufflefacenet,martinez2021benchmarking}   & 2.6                         & 577.5                   & 99.7                                                                 & 95.05                   & 88.50                                                                  & 96.9                                                                   & 97.3                                                                     & 92.3                                                                  & 94.3        \\ 
MobileFaceNet  \cite{martinez2021benchmarking}        & 2.0                         & 933                     & 99.7                                                                 & 95.2                    & 89.22                                                                  & 96.9                                                                   & 97.6                                                                     & 92.8                                                                  & 94.7 \\ 
\midrule
PocketNetM-256 \cite{boutros2022pocketnet}  & 1.75                        & 1099.15                 & 99.58                                                                & 95.63                   & 90.03                                                                  & \textbf{95.66}                                                                  & \textbf{97.17}                                                                    & 90.74                                                                 & 92.70  \\ 
PocketNetM-128 \cite{boutros2022pocketnet}  & 1.68                        & 1099.02                 & 99.65                                                                & 95.67                   & 90.00                                                                  & 95.07                                                                  & 96.78                                                                    & 90.63                                                                 & 92.63  \\ 
MixFaceNet-XS   \cite{boutros2021mixfacenets}       & 1.04                        & 161.9                   & 99.60                                                                & -                       & -                                                                      & -                                                                      & 95.85                                                                    & 88.48                                                                 & 90.73   \\ 
ShuffleMixFaceNet-XS  \cite{boutros2021mixfacenets}  & 1.04                        & 161.9                   & 99.53                                                                & -                       & -                                                                      & -                                                                      & 95.62                                                                    & 87.86                                                                 & 90.43   \\ 
MobileFaceNets   \cite{chen2018mobilefacenets}       & 0.99                        & 439.8                   & 99.55                                                                & -                       & -                                                                      & -                                                                      & 96.07                                                                    & -                                                                     & -  \\ 
PocketNetS-256 \cite{boutros2022pocketnet}  & 0.99                        & 587.24                  & 99.66                                                                & 95.50                   & 88.93                                                                  & 93.34                                                                  & 96.35                                                                    & 89.31                                                                 & 91.33   \\ 
PocketNetS-128 \cite{boutros2022pocketnet}  & 0.92                      & 587.11                  & 99.58                                                                & 95.48                   & 89.63                                                                  & 94.21                                                                  & 96.10                                                                    & 89.44                                                                 & 91.62    \\ 
ShuffleFaceNet 0.5x  \cite{martindez2019shufflefacenet}   & 0.5                         & 66.9                    & 99.23                                                                & -                       & -                                                                      & 92.59                                                                  & 93.22                                                                    & -                                                                     & - \\
\midrule

EdgeFace - S ($\gamma = 0.5$) \textbf{(ours)}   & 3.65         & 306.11            & 99.78            & \textbf{95.71}       & \textbf{92.56}      & 95.81      & 96.93         & \textbf{93.58}             & \textbf{95.63}          \\
EdgeFace - XS ($\gamma = 0.6$) \textbf{(ours)}   & 1.77         & 154            & \textbf{99.73}             & 95.28       & \textbf{91.82}      & 94.37      & 96.00         & \textbf{92.67}             & \textbf{94.85}         \\
\bottomrule
\end{tabular}}
\label{tab:verification}
\end{table*}

%% file: ablation_table.tex
\begin{table}
\caption{The performance of ablation of \textbf{XS} variant of \model on different face recognition datasets. The ablation is performance with respect to the $\gamma$ parameter. For top 5 rows, all values indicate verification accuracy (TAR) expressed as percentages; while for IJB-B and IJB-C datasets, the values refer to true positive rate (TPR) at false positive rate (FPR) of $1e-4$.}
\resizebox{\columnwidth}{!}{%
\begin{tabular}{l | r | r | r | r | r |r}
\toprule
\textbf{Dataset} & default & $\gamma=0.2$ & $\gamma=0.4$ & $\gamma=0.6$ & $\gamma=0.8$ & $\gamma=1.0$\\
\midrule
LFW     & 99.8  & 99.1  & 99.5  & 99.7  & 99.7  & 99.7 \\
CA-LFW  & 95.7  & 92.7  & 94.7  & 95.3  & 95.4  & 95.6 \\
CP-LFW  & 92.3  & 83.7  & 90.2  & 91.3  & 91.8  & 92.0 \\
AgeDB-30& 96.1  & 89.9  & 94.6  & 95.4  & 95.8  & 96.1 \\
CFP-FP  & 95.0  & 87.0  & 93.6  & 94.5  & 94.9  & 95.1 \\ \midrule
IJB-B    & 92.65 &  22.97 & 89.39& 92.24 & 92.80 & 93.24 \\
IJB-C    & 94.75 &  25.13 & 91.42& 94.28 & 95.01 & 95.02 \\ \midrule
\textbf{MPARAMS} &2.24       &0.73       &1.24       &1.77     &2.29       &2.81       \\
\textbf{MFLOPS} &196.9       &63.6       &107.9       &153.9     &198.4       & 244.4      \\

\bottomrule
\end{tabular}
}
\label{tab:ablation_perf}
\end{table}

%% file: competition_perf.tex
\begin{table*}

\setlength{\tabcolsep}{2.0pt}

\centering
\caption{The results from the IJCB 2023 Efficient Face Recognition Competition reproduced from the competition paper \cite{kolf2023efar}, along with the baselines. This includes details on FLOPs, model size, and the number of parameters. The achieved rank for each submission is specified for every dataset. The collective Borda count and rank across all verification benchmarks can be found in the Accuracy column. Rankings are also provided for FLOPs and model size. The final consolidated rank is a weighted Borda count considering the achieved accuracy (70\%), FLOPs ranking (15\%), and model size (15\%). Entries with 2-5M parameters are denoted as ``2-5 MP", while those with $<2$M parameters are labeled as ``2 MP".}
\resizebox{\textwidth}{!}{%
\begin{tabular}{c|c|cc|cc|cc|cc|cc|cc|cc|cc|cc|c|cc}
\toprule
\multicolumn{1}{l}{}            & \multicolumn{1}{l}{} & \multicolumn{4}{c|}{\textbf{Cross-Pose}}                                   & \multicolumn{4}{c}{\textbf{Cross-Age}}                                      & \multicolumn{2}{l}{}               & \multicolumn{2}{l}{}                   & \multicolumn{2}{l}{}                   & \multicolumn{2}{l}{}                & \multicolumn{2}{l}{}                     & \multicolumn{1}{l}{} & \multicolumn{2}{l}{}                   \\
\multicolumn{1}{c}{}            &                      & \multicolumn{2}{c|}{\textbf{CPLFW}} & \multicolumn{2}{c|}{\textbf{CFP-FP}} & \multicolumn{2}{c|}{\textbf{CALFW}} & \multicolumn{2}{c|}{\textbf{AgeDB30}} & \multicolumn{2}{c|}{\textbf{LFW}}  & \multicolumn{2}{c|}{\textbf{IJB-C}}    & \multicolumn{2}{c|}{\textbf{Accuracy}} & \multicolumn{2}{c|}{\textbf{FLOPS}} & \multicolumn{2}{c|}{\textbf{Model Size}} & \textbf{Params}      & \multicolumn{2}{c}{\textbf{Combined}}  \\
\textbf{Model}                  & \textbf{Category}    & \textbf{Acc. [\%]} & \textbf{Rank}  & \textbf{Acc. [\%]} & \textbf{Rank}   & \textbf{Acc. [\%]} & \textbf{Rank}  & \textbf{Acc. [\%]} & \textbf{Rank}    & \textbf{Acc. [\%]} & \textbf{Rank} & \textbf{TAR@$10^{-4}$} & \textbf{Rank} & \textbf{BC} & \textbf{Rank}            & \textbf{[M]} & \textbf{Rank}        & \textbf{[MB]} & \textbf{Rank}            & \textbf{[M]}         & \textbf{BC} & \textbf{Rank}            \\ 
\midrule

ResNet-100 ElasticFace (Cos+) \cite{boutros2022elasticface}  & Baseline             & 93.23              & -              & 98.73              & -               & 96.18              & -              & 98.28              & -                & 99.80              & -             & 96.65                  & -             & -           & -                        & 24211.778    &         -             & 261.22        & -                        & 65.2                 & -           & -                        \\ 

ResNet-100 ArcFace \cite{boutros2022elasticface, deng2019arcface}           & Baseline             & 92.08              & -              & 98.27              & -               & 95.45              & -              & 98.15              & -                & 99.82              & -             & 95.60                  & -             & -           & -                        & 24211.778    & -                    & 261.22        & -                        & 65.2                 & -           & -                        \\ 

ResNet-18 Q8-bit  \cite{DBLP:conf/icpr/BoutrosDK22}               & Baseline             & 89.48              & -              & 94.46              & -               & 95.72              & -              & 97.03              & -                & 99.63              & -             & 93.56                  & -             & -           & -                        & -            & -                    & 24.10         & -                        & 24.0                 & -           & -                        \\ 

ResNet-18 Q6-bit    \cite{DBLP:conf/icpr/BoutrosDK22}             & Baseline             & 88.37              & -              & 93.23              & -               & 95.58              & -              & 96.55              &          -        & 99.52              &       -        & 93.03                  & -             & -           & -                        & -            & -                    & 18.10         &       -                   & 24.0                 & -           & -                        \\ 
\midrule

MobileFaceNet Q8-bit   \cite{DBLP:conf/icpr/BoutrosDK22}          & Baseline             & 87.95              & -              & 91.40              & -               & 95.05              & -              & 95.47              & -                & 99.43              & -             & 90.57                  & -             & -           & -                        & -            & -                    & 1.10          & -                        & 1.1                  & -           & -                        \\ 

MobileFaceNet Q6-bit   \cite{DBLP:conf/icpr/BoutrosDK22}          & Baseline             & 84.57              & -              & 87.69              & -               & 93.30              & -              & 93.03              & -                & 98.87              & -             & 83.13                  & -             & -           & -                        & -            & -                    & 0.79          & -                        & 1.1                  & -           & -                        \\

PocketNetM-256 \cite{boutros2022pocketnet}                 & Baseline             & 90.03              & -              & 95.66              & -               & 95.63              & -              & 97.17              & -                & 99.58              & -             & 92.70                  & -             & -           & -                        & 1099.15      & -                    & 7.0           & -                        & 1.75                 & -           & -                        \\ 

PocketNetM-128 \cite{boutros2022pocketnet}                 & Baseline             & 90.00              & -              & 95.07              & -               & 95.67              & -              & 96.78              & -                & 99.65              & -             & 92.63                  & -             & -           & -                        & 1099.02      & -                    & 6.74          & -                        & 1.68                 & -           & -                        \\ 

\rowcolor{Gray}
\textbf{Idiap EdgeFace-XS($\gamma$=0.6)}  & 2 MP                 & 91.88              & \textbf{1}     & 94.46              & \textit{3}      & 95.25              & \textbf{1}     & 95.72              & \uline{2}        & 99.68              & \textbf{1}    & 94.78                  & \textbf{1}    & 39          & \textbf{1}               & 153.99       & 5                    & 7.17          & 7                        & 1.77                 & 5.0         & \textbf{1}               \\ 

\rowcolor{Gray}
Idiap EdgeFace-XXS-Q             & 2 MP                 & 89.65              & 5              & 93.11              & 5               & 94.68              & 4              & 93.77              & 4                & 99.50              & 4             & 92.97                  & 4             & 22          & 4                        & 94.72        & \textit{3}           & 1.73          & \uline{2}                & 1.24                 & 3.92        & 4                        \\ 

MobileNet$_\text{V2-visteam}$    & 2 MP                 & 82.90              & 8              & 89.39              & 7               & 88.63              & 6              & 83.65              & 7                & 98.58              & 6             & 51.60                  & 7             & 7           & 7                        & 86.20        & \uline{2}            & 3.38          & \textit{3}               & 1.70                 & 2.17        & 6                        \\ 

SAM-MFaceNet eHWS V1             & 2 MP                 & 91.35              & \uline{2}      & 95.01              & \textbf{1}      & 95.10              & \uline{2}      & 95.57              & \textit{3}       & 99.55              & \textit{3}    & 93.07                  & \uline{2}     & 35          & \uline{2}                & 236.75       & 7                    & 4.4           & 5                        & 1.10                 & 4.68        & \uline{2}                \\ 

SAM-MFaceNet eHWS V2             & 2 MP                 & 91.28              & \textit{3}     & 94.73              & \uline{2}       & 94.90              & \textit{3}     & 95.72              & \textbf{1}       & 99.65              & \uline{2}     & 93.06                  & \textit{3}    & 33          & \textit{3}               & 236.75       & 6                    & 4.4           & 6                        & 1.10                 & 4.45        & 3                        \\ 

SQ-HH                            & 2 MP                 & 84.13              & 6              & 91.60              & 6               & 87.17              & 7              & 84.28              & 6                & 98.07              & 7             & 63.36                  & 6             & 10          & 6                        & 1399.39      & 8                    & 4.55          & 8                        & 1.20                 & 1.47        & 8                        \\ 

ShuffleNetv2x0.5                 & 2 MP                 & 83.48              & 7              & 87.76              & 8               & 86.00              & 8              & 80.33              & 8                & 97.72              & 8             & 38.57                  & 8             & 1           & 8                        & 17.14        & \textbf{1}           & 0.77          & \textbf{1}               & 0.17                 & 1.92        & 7                        \\ 

ShuffleNetv2x1.5                 & 2 MP                 & 89.73              & 4              & 93.44              & 4               & 91.08              & 5              & 88.78              & 5                & 98.95              & 5             & 77.11                  & 5             & 20          & 5                        & 147.21       & 4                    & 7.90          & 4                        & 1.99                 & 2.78        & 5    \\   

\midrule

VarGFaceNet  \cite{yan2019vargfacenet}                   & Baseline             & 88.55              & -              & 98.50              & -               & 95.15              & -              & 98.15              & -                & 99.85              & -             & 94.70                  & -             & -           & -                        & 1022         & -                    & 20.0          & -                        & 5.0                  & -           & -                        \\ 

MobileFaceNetV1 \cite{chen2018mobilefacenets}                & Baseline             & 87.17              & -              & 95.80              & -               & 94.47              & -              & 96.40              & -                & 99.40              & -             & 93.90                  & -             & -           & -                        & 1100         & -                    & 13.6          & -                        & 3.4                  & -           & -                        \\ 

MixFaceNet-M \cite{boutros2021mixfacenets}                   & Baseline             & -                  & -              & -                  & -               & -                  & -              & 97.05              & -                & 99.68              & -             & 93.42                  & -             & -           & -                        & 626.1        & -                    & 15.8          & -                        & 3.95                 & -           & -                        \\ 

MixFaceNet-S \cite{boutros2021mixfacenets}                   & Baseline             & -                  & -              & -                  & -               & -                  & -              & 96.63              & -                & 99.60              & -             & 92.30                  & -             & -           & -                        & 451.7        & -                    & 12.28         & -                        & 3.07                 & -           & -                        \\ 

MixFaceNet-XS \cite{boutros2021mixfacenets}                  & Baseline             & -                  & -              & -                  & -               & -                  & -              & 95.85              & -                & 99.60              & -             & 90.73                  & -             & -           & -                        & 161.9        & -                    & 4.16          & -                        & 1.04                 & -           & -                        \\ 

ShuffleMixFaceNet-M \cite{boutros2021mixfacenets}            & Baseline             & -                  & -              & -                  & -               & -                  & -              & 96.98              & -                & 99.60              & -             & 91.47                  & -             & -           & -                        & 626.1        & -                    & 15.8          & -                        & 3.95                 & -           & -                        \\ 

ShuffleMixFaceNet-S \cite{boutros2021mixfacenets}            & Baseline             & -                  & -              & -                  & -               & -                  & -              & 97.05              & -                & 99.58              & -             & 93.08                  & -             & -           & -                        & 451.7        & -                    & 12.28         & -                        & 3.07                 & -           & -                        \\ 

ShuffleFaceNet 1.5x \cite{martindez2019shufflefacenet}             & Baseline             & 88.50              & -              & 97.26              & -               & 95.05              & -              & 97.32              & -                & 99.67              & -             & 94.30                  & -             & -           & -                        & 577.5        & -                    & 10.5          & -                        & 2.6                  & -           & -                        \\ 

MobileFaceNet \cite{chen2018mobilefacenets}                  & Baseline             & 89.22              & -              & 96.90              & -               & 95.20              & -              & 97.60              & -                & 99.70              & -             & 94.70                  & -             & -           & -                        & 933          & -                    & 4.50          & -                        & 2.0                  & -           & -                        \\

EfficientNet$_\text{b0-visteam}$ & 2-5 MP               & 87.58              & 9              & 91.19              & 9               & 93.35              & 7              & 90.45              & 7                & 99.15              & 8             & 85.04                  & 7             & 7           & 8                        & 212.50       & \textit{3}           & 9.18          & 7                        & 4.60                 & 1.87        & 8                        \\ 

GhostFaceNetV1-1 KU              & 2-5 MP               & 91.70              & 4              & 95.00              & 5               & 95.77              & \textbf{1}     & 97.20              & \textbf{1}       & 99.62              & \textit{3}    & 94.93                  & \uline{2}     & 37          & \uline{2}                & 215.65       & 4                    & 8.17          & 4                        & 4.09                 & 5.52        & \textbf{1}               \\ 

GhostFaceNetV1-2 KU              & 2-5 MP               & 90.03              & 7              & 93.30              & 7               & 95.72              & \uline{2}      & 97.08              & \uline{2}        & 99.72              & \uline{2}     & 94.06                  & 4             & 29          & 5                        & 60.29        & \textbf{1}           & 8.07          & \uline{2}                & 4.06                 & 5.33        & \uline{2}                \\ 

\rowcolor{Gray}
\textbf{Idiap EdgeFace-S($\gamma$=0.5)}   & 2-5 MP               & 92.22              & \textit{3}     & 95.67              & \textit{3}      & 95.62              & \textit{3}     & 96.98              & \textit{3}       & 99.78              & \textbf{1}    & 95.63                  & \textbf{1}    & 39          & \textbf{1}               & 306.11       & 5                    & 14.69         & 8                        & 3.65                 & 5.15        & \textit{3}               \\ 

\rowcolor{Gray}
Idiap EdgeFace-XS-Q              & 2-5 MP               & 90.92              & 5              & 94.26              & 6               & 95.03              & 6              & 95.22              & 6                & 99.50              & 6             & 94.40                  & \textit{3}    & 22          & 6                        & 196.91       & \uline{2}            & 2.99          & \textbf{1}               & 2.24                 & 4.52        & 4                        \\ 

MB2-HH                           & 2-5 MP               & 90.65              & 6              & 95.13              & 4               & 91.43              & 8              & 90.08              & 8                & 99.32              & 7             & 79.86                  & 9             & 12          & 7                        & 741.67       & 9                    & 8.15          & \textit{3}               & 2.20                 & 2.15        & 7                        \\ 

Modified-MobileFaceNet V1        & 2-5 MP               & 92.42              & \textbf{1}     & 95.97              & \uline{2}       & 95.15              & 5              & 95.77              & 5                & 99.52              & 5             & 93.99                  & 5             & 31          & 4                        & 456.89       & 8                    & 8.4           & 6                        & 2.10                 & 4.22        & 6                        \\ 

Modified-MobileFaceNet V2        & 2-5 MP               & 92.23              & \uline{2}      & 96.11              & \textbf{1}      & 95.15              & 4              & 95.88              & 4                & 99.58              & 4             & 93.95                  & 6             & 32          & \textit{3}               & 456.89       & 7                    & 8.4           & 5                        & 2.10                 & 4.33        & 5                        \\ 

ShuffleNetv2x2.0                 & 2-5 MP               & 89.27              & 8              & 92.71              & 8               & 90.88              & 9              & 88.08              & 9                & 99.03              & 9             & 80.92                  & 8             & 3           & 9                        & 310.92       & 6                    & 20.00         & 9                        & 4.97                 & 0.65        & 9                        \\ 
\bottomrule
\end{tabular}
}
\label{tab:competition_table}
\end{table*}